\documentclass[10pt,twocolumn,letterpaper]{article}

 \usepackage{cvpr}              
   
\usepackage[dvipsnames]{xcolor}
\usepackage{cite}
\usepackage{amsmath,amssymb,amsfonts}
\usepackage{algorithmic}
\usepackage{graphicx}
\usepackage{tabularx}
\usepackage{booktabs}
\usepackage{textcomp}
\usepackage{xcolor}
\usepackage{multirow}
\usepackage{comment}
\usepackage[normalem]{ulem}

\definecolor{cvprblue}{rgb}{0.21,0.49,0.74}
\usepackage[pagebackref,breaklinks,colorlinks,citecolor=cvprblue]{hyperref}

\def\paperID{7} 
\def\confName{CVPR}
\def\confYear{2024}

\title{Energy-Efficient Uncertainty-Aware Biomass \\ Composition Prediction at the Edge}
\author{Muhammad Zawish$^{\dagger}$, Paul Albert$^{\ddagger}$, Flavio Esposito$^{\mathsection}$, Steven Davy$^{\P}$, and Lizy Abraham$^{\dagger}$\\ \\
$^{\dagger}$Walton Institute, South East Technological University, Ireland\\
$^{\ddagger}$Center for Augmented Reasoning, Australian Institute of Machine Learning, Australia\\
$^{\mathsection}$Department of Computer Science, Saint Louis University, USA\\
$^{\P}$Centre for Sustainable Digital Technologies at Technological University Dublin, Ireland\\
{\tt\small \{muhammad.zawish\}@waltoninstitute.ie}
}
\begin{document}
\maketitle
\begin{abstract}
\label{sec:abstract}
Clover fixates nitrogen from the atmosphere to the ground, making grass-clover mixtures highly desirable to reduce external nitrogen fertilization. Herbage containing clover additionally promotes higher food intake, resulting in higher milk production. 
Herbage probing however remains largely unused as it requires a time-intensive manual laboratory analysis. Without this information, farmers are unable to perform localized clover sowing or take targeted fertilization decisions. 
Deep learning algorithms have been proposed with the goal to estimate the dry biomass composition from images of the grass directly in the fields. The energy-intensive nature of deep learning however limits deployment to practical edge devices such as smartphones. 
This paper proposes to fill this gap by applying filter pruning to reduce the energy requirement of existing deep learning solutions. 
We report that although pruned networks are accurate on controlled, high-quality images of the grass, they struggle to generalize to real-world smartphone images that are blurry or taken from challenging angles. We address this challenge by training filter-pruned models using a variance attenuation loss so they can predict the uncertainty of their predictions. When the uncertainty exceeds a threshold, we re-infer using a more accurate unpruned model. This hybrid approach allows us to reduce energy-consumption while retaining a high accuracy. We evaluate our algorithm on two datasets: the GrassClover and the Irish clover using an NVIDIA Jetson Nano edge device. We find that we reduce energy reduction with respect to state-of-the-art solutions by 50\% on average with only 4\% accuracy loss. 
\end{abstract}    
\begin{figure}
\centering
\includegraphics[width=0.4\textwidth]{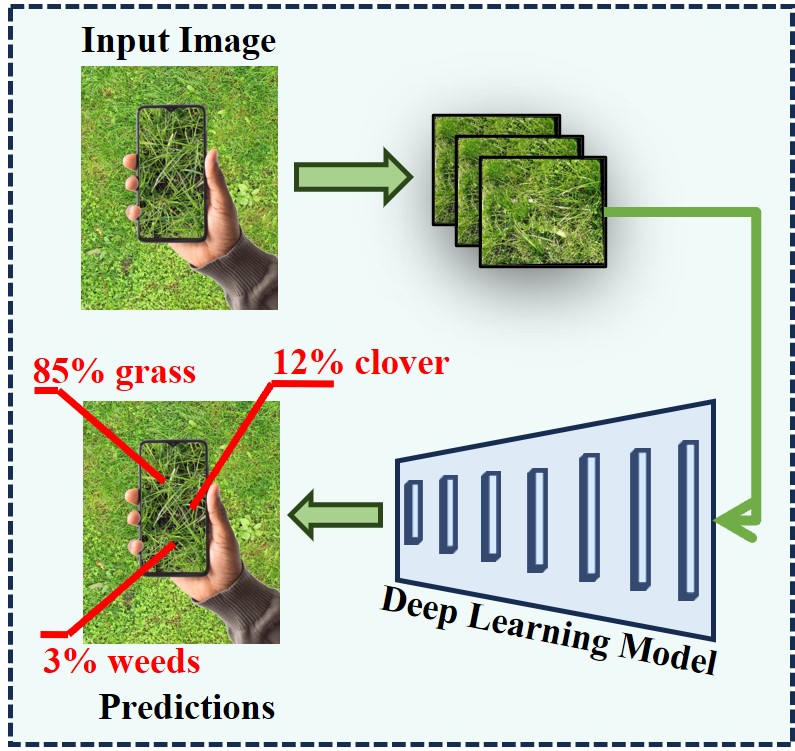}
\caption{A typical workflow of deep learning based biomass estimation from phone / camera images.}
\label{Fig1}
\end{figure}
\section{Introduction}
\label{sec:intro}
Evaluating the dry biomass composition of herbage delivers insights that can in turn be used by farmers to improve the quality of their grass. The usual biomass composition process requires cutting  grass on a delimited surface in the field. The grass sample is then taken back to a laboratory where it is dried in a low-temperature oven before being manually separated in its components~\cite{2018_ICPA_grasscloverfirst,2021_EGF_Irishdataset}. This process is constrained by time and human labor which hinders its practical application. Computer vision techniques have the potential to reduce this laborious process by predicting dry biomass composition from in situ images. Using image processing via deep learning is also non-destructive for the soil, and a faster method to estimate the biomass can facilitate the adoption of fine-grained herbage probing. Figure~\ref{Fig1} illustrates such a process when images are collected with smartphones.

Computer vision systems have been previously developed on a variety of images including high-resolution camera systems mounted on All Terrain Vehicles (ATVs)~\cite{2019_CVPRW_grasscloverdataset} or on manually operated camera tripods~\cite{2021_EGF_Irishdataset}. Other more practical approaches include using phones~\cite{2022_EGF_Irishunsuppre} or drone images~\cite{2022_CVPRW_IrishSuperRes}. These systems, however, were not designed to address the limited resources of the targeted edge devices, especially in terms of the energy consumption of large neural networks that can limit the granularity of biomass estimation on battery-powered devices. 

Model compression, particularly structured (filter-based) pruning has proved to be a viable solution for reducing model size and computations~\cite{anwar2017structured}. It typically measures the saliency of filters in $conv$ layers and prunes them according to a predefined threshold. The idea of mapping saliency with the loss function makes these approaches dependent on training data, hence limiting their practicability on specific computer vision tasks, such as image classification. Recent work has shown that even without measuring filter saliency, it is possible to achieve the same level of performance with random pruning~\cite{mittal2019studying}. Furthermore, pruning approaches in the literature mostly focus on pruning post-training to effectively cut down on inference costs, while inevitably increasing training and re-training overhead. Recently, a new method has emerged to prune models at initialization (before training) \cite{frankle2018lottery, zawish2024}, and show similar performances to the post-training pruning approaches. 
We find energy-aware pruning at \textit{initialization} attractive for biomass estimation because although prior studies demonstrate the accuracy of deep learning models to solve the task, these large models are intended for deployment on energy-constrained edge devices like smartphones. Energy-aware pruning indeed offers the potential for large energy savings, however, end-users require reliable results to facilitate technology adoption. 
We aim in this paper to answer the need for low energy consumption while retaining biomass prediction accuracy.

When training state-of-the-art biomass composition estimation models~\cite{2022_CVPRW_IrishSuperRes} using initialization-pruned models, we observe a significant decrease in accuracy concentrated in a population of \textit{harder} images. These images are clover-heavy or were taken with hand-held smartphones, leading to lower image quality and complicating the biomass estimation task. To preserve a high prediction accuracy, we propose to identify these hard images at test-time so that they can be re-predicted on using an unpruned model, inherently more robust to these challenging examples. We achieve this by training the pruned models to predict a normally distributed probability distribution over biomass values rather than predicting a single scalar biomass value. The probability distribution is characterized by a mean $\mu$ and a variance $\sigma$. We find that $\sigma$ is positively correlated with the prediction error of the pruned network on the hard images. At test-time and given the prediction of the pruned network on a target image, if we detect that $\sigma$ exceeds a threshold, i.e., the prediction is under-confident, we offset such predictions using the unpruned network to increase accuracy. In the case where $\sigma$ is small enough, $\mu$ is used as the energy-efficient and reliable predictions. In effect, during the deployment of our algorithm on edge devices, we can confidently save energy on simpler images and call for the more robust yet energy-hungry unpruned model for the detected challenging images.

Consequently, the proposed approach conserves better accuracy than the pruned model while yielding higher energy efficiency than the unpruned model.\\
In particular, in this paper, we present the following key contributions:
\begin{itemize}
\item We perform stochastic filter pruning at initialization as in~\cite{zawish2024}, iteratively removing filters until the pruning quota is met, with high-energy $conv$ layers being more likely to be pruned. We consequently present an evaluation of the detrimental effect of pruning rate on prediction accuracy.
\item We introduce a uncertainty-guided approach to accommodate performance loss in pruned models. A high variance $\sigma$ of the biomass  probability distribution predicted by the pruned models informs the algorithm on the subset of harder images that need to be more accurately predicted by the unpruned model.
\item We compare with uncompressed deep learning baselines from the state-of-the-art~\cite{2022_CVPRW_IrishSuperRes} using two network architectures~\cite{2016_CVPR_ResNet,2015_ICLR_VGG} on two grass biomass estimation datasets~\cite{2019_CVPRW_grasscloverdataset,2021_EGF_Irishdataset} where we show we can maintain a comparable accuracy while decreasing energy usage. 
\item We measure the real-world energy efficiency of our algorithm on a NVIDIA Jetson Nano, a resource-constrained edge device where we realise a $40\%$ to $60\%$ energy consumption decrease depending on the network and dataset. We hope that this large decrease in energy consumption further motivates the deployment of biomass estimation algorithms to edge devices without compromising much on accuracy. 

\end{itemize}
\section{Related Work}
\label{sec:related}
This section presents into existing literature on dry biomass estimation from images and model compression. While significant research exists on model pruning and accurate biomass estimation using deep learning, we notice a notable gap in integrating both for biomass estimation at the edge.

\subsection{Grass and clover dry biomass estimation}

We first detail datasets and research relevant to grass and clover dry biomass estimation using computer vision.
The GrassClover dataset~\cite{2018_ICPA_grasscloverfirst,2019_CVPRW_grasscloverdataset} proposes to predict the dry biomass composition of grass from images taken from a high resolution camera mounted on an ATV. $179$ images of grass-clover mixtures are collected, oven-dried and manually labeled to obtain the ground-truth. The baseline biomass estimation proposed by the authors is a two stage approach where a large semantic segmentation model trained on synthetic data is used to predict the species of each pixel in the grass canopy. Once the canopy composition is estimated, a linear regression infers the dry biomass composition. In a further iteration~\cite{2021_SENSORS_skovsen2}, the authors utilize a style transfer generative network~\cite{2020_IET_glstylenet} to generate data augmentations on the collected images to simulate different weather conditions. This effectively increases the variety of the training dataset and enables more robust predictions.
Because the GrassClover dataset contains a set of partially annotated images, researchers have proposed to use a one stage regression from image to dry biomass using a mean-imputation approach using a VGG16~\cite{2015_ICLR_VGG} neural network~\cite{2021_EGF_Irishdataset}. Other solutions utilize the semantic segmentation into linear separation baseline to approximately label unlabeled images so they can be introduced in the training set~\cite{2021_ICCVW_semisupmine}, adapt the MixMatch semi-supervised classification algorithm~\cite{2019_NeurIPS_mixmatch} to the regression problem~\cite{2022_CVPRW_IrishSuperRes} or to use them for self-supervised pre-training~\cite{2022_EGF_Irishunsuppre}. In addition to the GrassClover dataset, the IrishClover~\cite{2021_EGF_Irishdataset} dataset consists of $424$ training images captured using cameras mounted on tripods and proposes a similar challenge of dry biomass estimation with added difficulty of unbalanced species representation and the presence of handheld phone images in the test set.

\subsection{Lightweight neural networks for edge AI in agriculture}
Generally, majority of the deep learning-based agricultural applications discussed above are cloud-centric. This means that during the test-time, the incoming data from mobile edge devices would be offloaded to a remote cloud server to perform the prediction process. For example, authors in \cite{apolo2020cloud} proposed a cloud-based AI solution for predicting biomass yield maps from apple orchards using images captured through UAV. Cloud has remained a viable solution for these tasks, since deep learning models are computationally-intensive and consume significant energy and storage capacity. However, it requires continuous and stable internet connection to approach cloud for deep learning tasks, which is hard to realize in remote farming environments. Recently, researchers have proposed to process the data directly on the edge devices to achieve fast responses while ensuring data privacy. For example, authors in \cite{der2020real} propose to directly classify the crop types using a deep learning model deployed on an embedded board over a UAV. Similarly, Camargo et al. \cite{de2021optimized} customized a lightweight version of ResNet model to fit on an NVIDIA Jetson Xavier board for automatic weed mapping. However, these customized lightweight approaches do not provide flexible and scalable models, since these models are only developed for specific tasks. For these edge applications, model compression approaches have recently emerged, particularly pruning to achieve compression from coarse (low compression) to fine (high compression) scales \cite{han2015deep, anwar2017structured, frankle2018lottery, zawish2024}. In this paper, we implement filter pruning as a type of model compression to realize lightweight models for biomass estimation on different scales. We also show how to achieve energy efficiency on a NVIDIA Jetson Nano edge device while preserving accuracy using the best of both pruned and unpruned models.

\begin{figure*}
\includegraphics[width=1\textwidth]{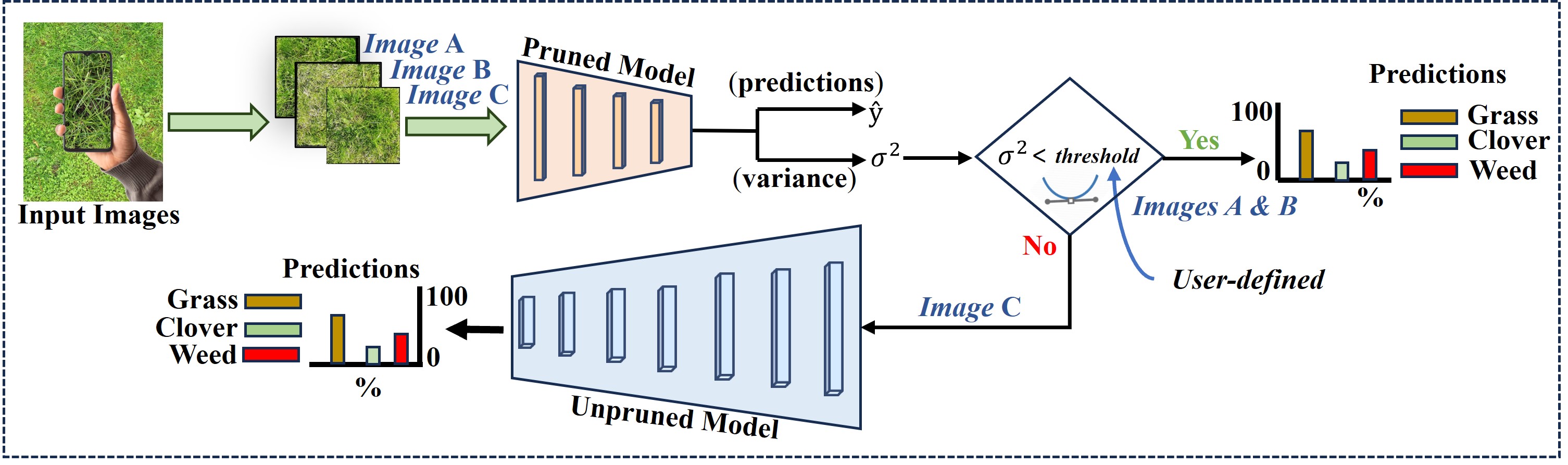}
\caption{Our proposed framework based on variance-guided energy efficient deep learning for biomass estimation on edge.}
\centering\label{Fig2}
\end{figure*}
\section{Methodology}
In this section, we first motivate the idea applying model compression to biomass composition estimation algorithms. We then show how to circumvent the performance loss of pruned models using our variance-guided approach. 
\subsection{Motivation}
Biomass estimation using deep learning is generally carried out on a high-end remote GPU / CPU cloud servers~\cite{2021_ICCVW_semisupmine}. This is because SoTA deep learning models are resource-intensive in nature, requiring significant amounts of storage and power consumption. Using cloud-based inference implies that input images captured through mobile phones need to be offloaded -- via a wireless network -- to the cloud. This, however, is not considered reliable for making real-time predictions, since continuous and stable internet connectivity is not guaranteed in remote agricultural farms. Moreover, offloading raw images from farms to cloud poses privacy concerns, which can be avoided by leveraging on-device processing capabilities of mobile edge devices. Numerous research efforts have been in place to accommodate deep models directly on mobile phones. Model compression, particularly filter pruning, has been widely used to enable on-device AI for several computer vision applications but at the cost of slight reduction in accuracy \cite{han2015deep}. Moreover, most of the compression approaches are evaluated on image classification which raises questions about their functionality for other tasks. This paper proposes to apply model pruning at initialization on a vision-based regression task for biomass estimation. In the subsequent section, we further discuss the pruning approach.

\subsection{Pruning at Initialization}
Pruning in neural networks is typically of two types; \textit{i) structured} and \textit{ii) unstructured}. Early works on pruning favoured unstructured approach due to its capacity to provide high compression ratios at a finer granularity such as individual weight-level~\cite{han2015deep}. However, recent works have practically proven that unstructured sparsity merely achieves realistic acceleration due to (weight) index storage overhead, and requires specialized libraries for matrix operation \cite{ma2021non}. In contrast, removing whole filters/kernels from \textit{convolution} layers is a structured form of pruning since it preserves the structure of the network. Therefore, filter pruning has gained attention in reducing the model size and computations while being implementation-friendly \cite{anwar2017structured}. For this reason, we opt for filter pruning to induce structured sparsity in the baseline (unpruned) models.

Traditionally, filter pruning observes a three-stage procedure: \textit{-- train, prune and re-train}~\cite{anwar2017structured}. Although this setting is widely adopted, it introduces computational overhead characterized by multiple rounds of training and re-training. An alternative paradigm has however emerged namely the -- Lottery Ticket Pruning, where pruning is conducted prior to training~\cite{frankle2018lottery}. In the case of lottery ticket pruning, sparse models are created at initialization from a dense (unpruned) model without incurring a pre-training overhead and while realising similar accuracy guarantees~\cite{zawish2024}. Additionally, during the pruning phase, the saliency of filters is assessed using predefined criteria such as the \textit{l}\textsubscript{1}-norm \cite{l1norm}, enabling the removal of filters with saliency scores below a certain threshold. In contrast, ``\textit{randomly}" pruning filters has recently been proposed to also match the performance of saliency-based approaches, without introducing unnecessary computations of measuring saliency \cite{mittal2019studying}. Nevertheless, it is important to note that the success of random pruning lies in careful selection of candidate \textit{convolution} layers for filter pruning \cite{zawish2024,mittal2019studying}. 

\subsection{Energy-driven layer selection}
Typically, in the majority of pruning scenarios, networks are pruned \textit{uniformly} where each \textit{convolution} layer end up with the same proportion of removed filters. This not only results in a larger loss of performance but also leads to an unfair pruning regime, since every layer exhibits different level of underlying complexity. Inspired from~\cite{zawish2024}, we select the candidate layer for random filter-pruning in an energy-driven manner, until the desired compression scale ($\varepsilon$) is met. In particular, during each pruning iteration, the probability ${P_k}$ of $conv$ layer $k$ to be selected is determined by its relative energy consumption as shown in Eq. \ref{eq5}:

\begin{equation} \label{eq5}
    \mathit {P_k}=\frac{E_k}{\sum\limits_{k\in K} E_k},
\end{equation}

where $(E_k)$ is total energy requirement of a layer $k$, which is sum of computational energy $(E^{FLOPs}_{k})$ and energy for memory access $(E^{Access}_{k})$. In this way, layers are penalized (pruned) in order of their contribution to the overall energy consumption of the model. This approach was shown in~\cite{zawish2024} to not only restricts the layer collapse but also keeps sufficient capacity within the model to capture important patterns from the data. Below we elaborate further on the individual terms $E^{FLOPs}_{k}$ and $E^{Access}_{k}$.

\textbf{Energy for computations.}
The computational energy $(E^{FLOPs}_{k})$ requirement within a \textit{convolution} layer emerges from its underlying matrix multiplications and additions. These arithmetic operations are represented theoretically as Floating-point operations (FLOPs). For a \textit{convolution} layer $k$, the number of FLOPs can be captured using Eq. \ref{eq1}:

\begin{equation} \label{eq1}
   	k_{flops}= {[C_{in}\times (\Omega)^2 \times  C_{out}\times S_{out}]},
\end{equation}
then, the energy consumption for performing these computations can be modelled using Eq. \ref{eq2}: 

\begin{equation}\label{eq2}
E^{FLOPs}_{k} = k_{flops} \times A,
\end{equation}
where $C_{in}$ and $C_{out}$ represent the input and output channels or filters, respectively. $\Omega^2$ indicates the filter dimension, $S_{out}$ is the feature size of the output layer, and $A$ denotes the energy required for a single FLOP, i.e., 2.3pJ \cite{han2016eie}. 

\textbf{Energy for memory access.}
Accessing data stored in the memory also contributes to significant amount of energy consumption. For a \textit{convolution} layer $k$, the amount of memory required can be formulated using Eq. \ref{eq3}: 
\begin{equation}\label{eq3}
    k_{mem} = {[(k_{params}\times 3) + \{(S_{out} \times  C_{out}^{k}) \times 2 \}]}\times 4,
\end{equation}
then, the energy consumption for accessing this amount of data can be modelled using Eq. \ref{eq4}: 

\begin{equation}\label{eq4}
E^{Access}_{k} = k_{mem} \times B,
\end{equation}
where $k_{params}$ are the number of parameters, and $B$ denotes the energy required to access 1MB of data from DRAM i.e., 640pJ \cite{han2016eie}. In Eq. \ref{eq3}, the constant 3 accounts for gradients and momentum variables apart from parameters, and 2 accounts for storing the activations and their errors during the backpropagation. Finally, 4 refers to number of bytes required to store in a 32-bit system.

\subsection{Uncertainty and biomass estimation}
Although model compression reduces the energy consumption of neural networks, this cannot be attained without sacrificing on accuracy. In the case of biomass composition estimation from phone images, we specifically observe that the loss in accuracy is constrained to specific input images that become harder to predict for the less generic energy-constrained models. Therefore, we propose to identify unreliable predictions of the pruned model on the fly and to refine them with accurate predictions from the unpruned model at test time. This enables high energy efficiency by using the energy-pruned model on most images while retaining high accuracy using the unprunned model on the most challenging ones.

\begin{figure}
\includegraphics[width=0.5\textwidth]
{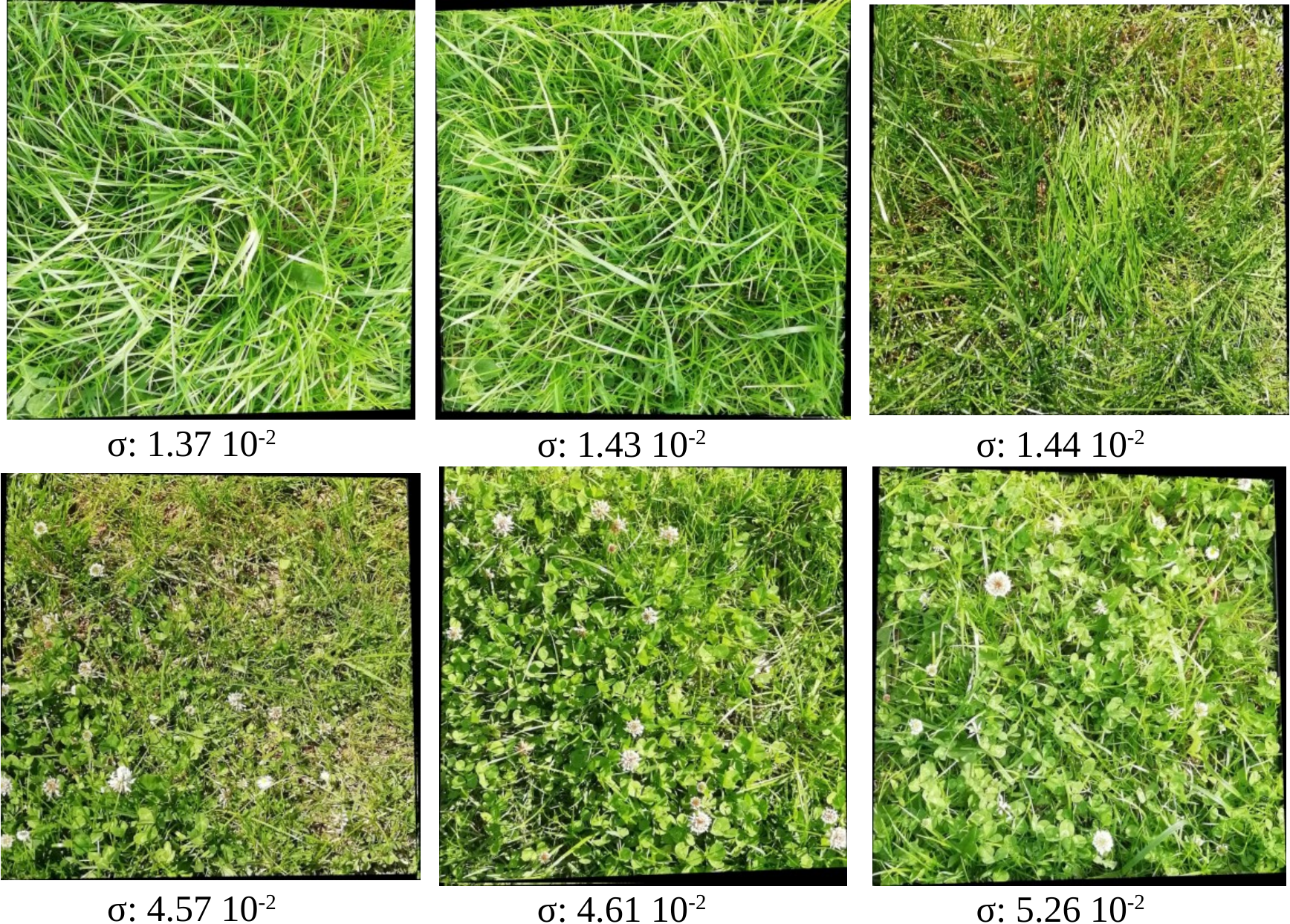}
\caption{Examples of predicted high and low variance images from the Irish dataset, ResNet18 $80\%$ pruned.}
\label{irishvar}
\end{figure}

\begin{figure}
\includegraphics[width=0.5\textwidth]
{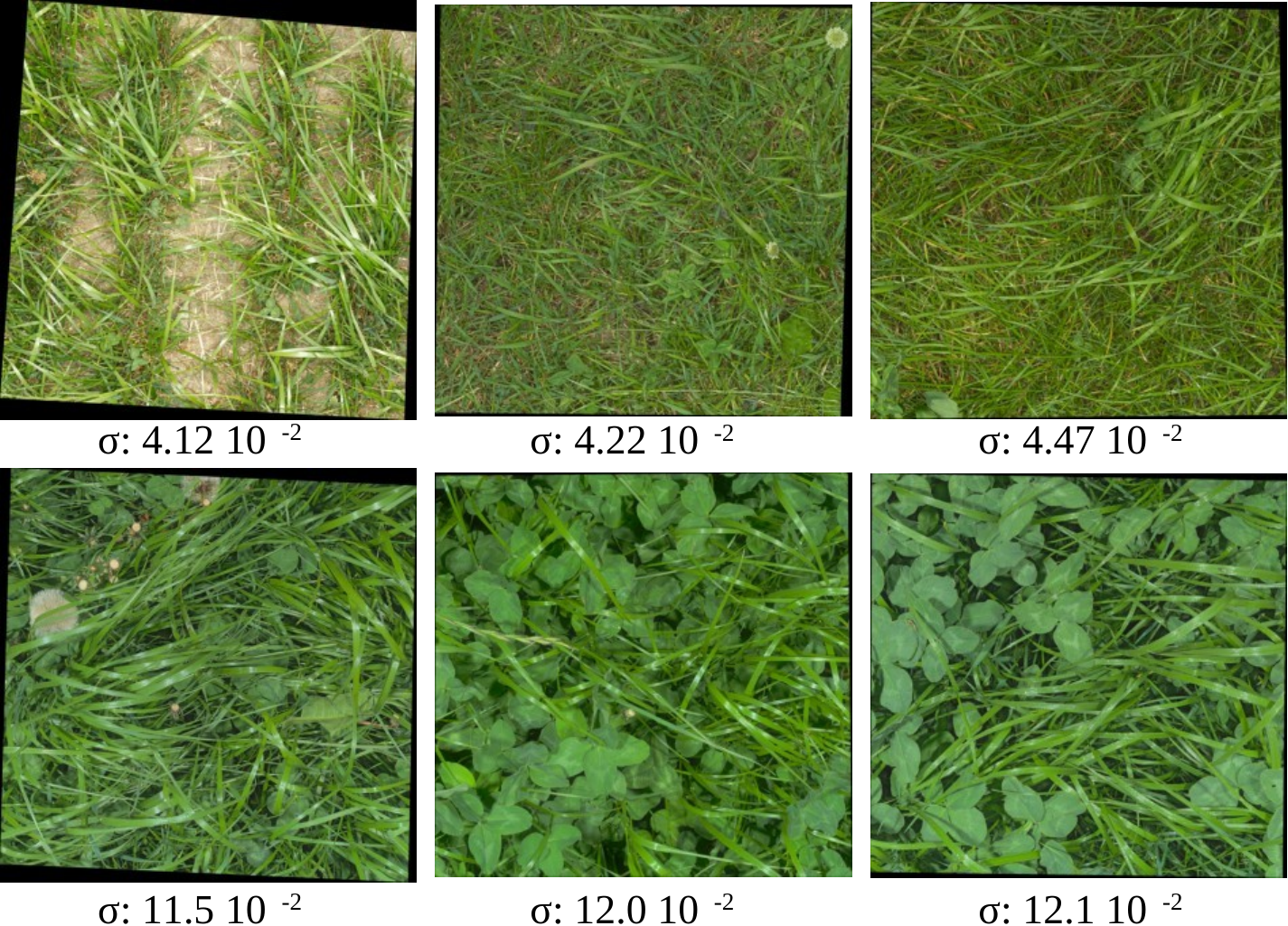}
\caption{Examples of predicted high and low variance images from the GrassClover dataset, ResNet18 $80\%$ pruned.}
\label{fig:danishvar}
\end{figure}

\begin{figure}[t]
\includegraphics[width=0.5\textwidth]
{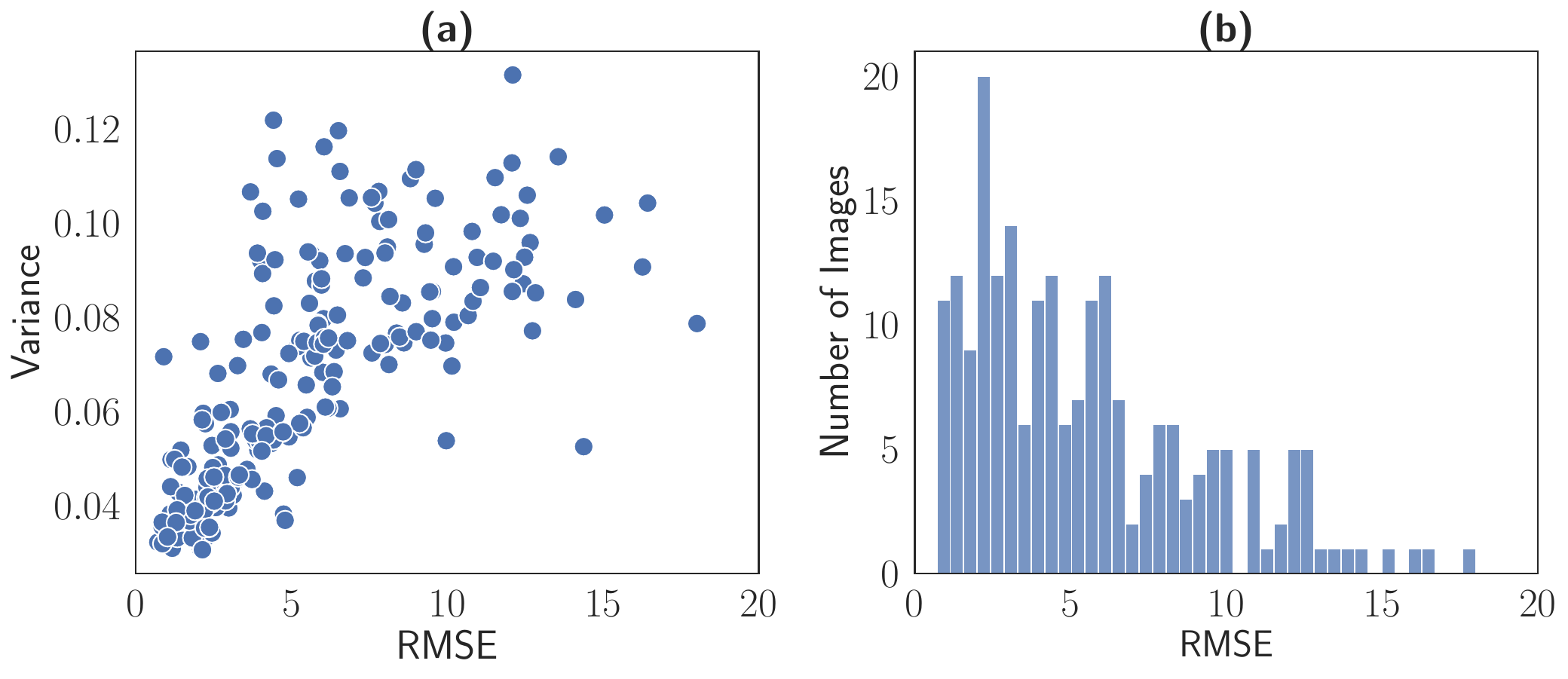}
\caption{(a) showing the correlation between RMSE and Variance, (b) showing the RMSE distribution over test set of Irish clover dataset for VGG-16.}
\label{Fig3}
\end{figure}

\begin{figure}[t]
\includegraphics[width=0.5\textwidth]
{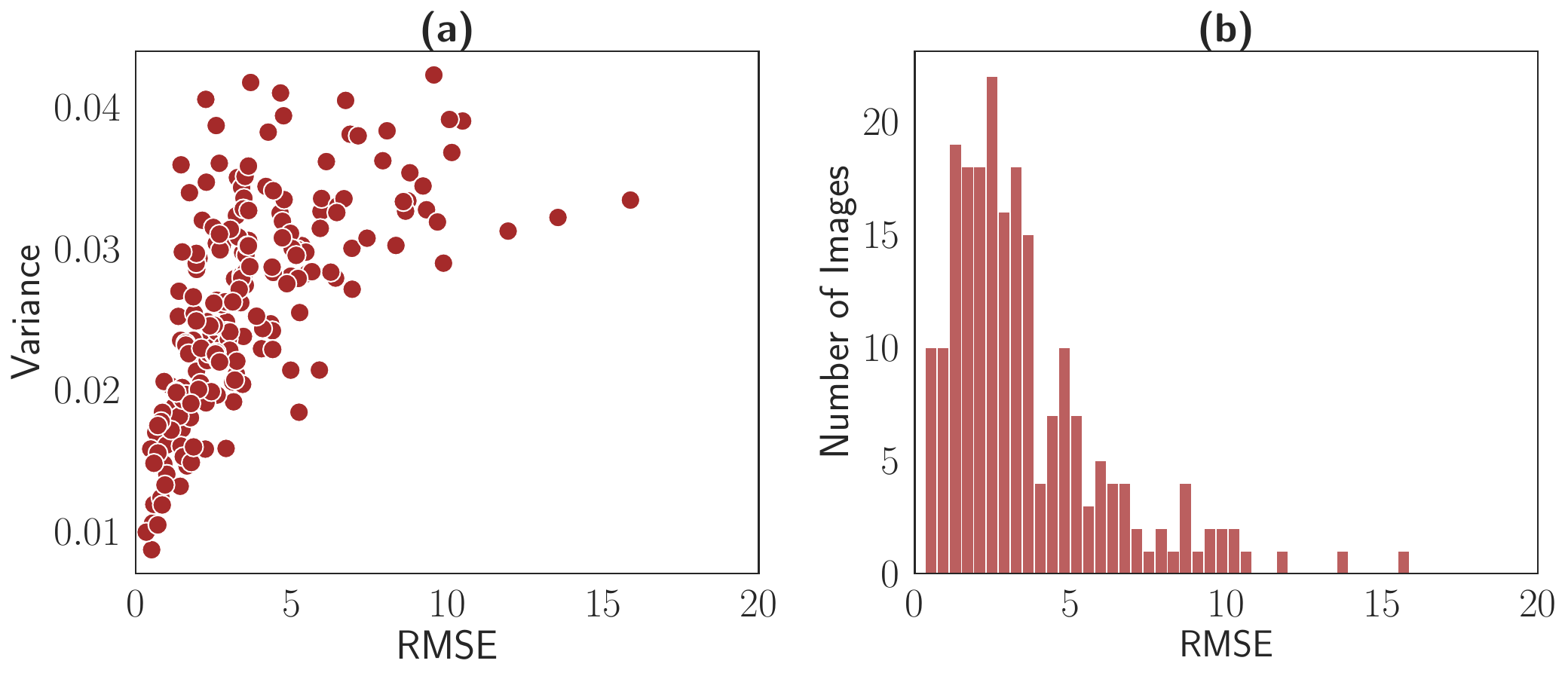}
\caption{(a) showing the correlation between RMSE and Variance, (b) showing the RMSE distribution over test set of Irish clover dataset for ResNet-18.}
\label{Fig4}
\end{figure}

\textbf{Variance attenuation loss.}
Since Figure~\ref{Fig3}(b) and Figure~\ref{Fig4}(b) evidence that the errors made by the pruned models are concentrated in a few images, we aim to detect these harder samples. 
Although prediction confidence is often linked to prediction accuracy, estimating confidence in the context of regression is non evident. We propose to train our network to expect the ground-truth biomass target $y$ for an input image $x$ to not be a scalar biomass value but to be an observation from a normal distribution, centered around a mean $y(x)$ and corrupted by some input dependant noise $\sigma(x)$: $ y \sim \mathcal{N}(\mu(x), \sigma(x))$. $\mu(x)$ here is the most likely value for $y$ (mean of the distribution) predicted by the network from the input image $x$.

We train our network to maximize the following likelihood function over the dataset $\mathcal{D} = \{x_i, y_i\}_{i=1}^N$ 
\begin{equation}\label{eq6}
\prod_{i=1}^N \frac{1}{\sigma(x_i)\sqrt{2\pi}}e^{-\frac{1}{2}\left(\frac{y_i-\mu(x_i)}{\sigma(x_i)}\right)^2}
\end{equation}
equivalent to
\begin{equation}\label{eq7}
 \frac{1}{(\sqrt{2\pi})^2}e^{-\frac{1}{2}\sum_{i=1}^N2\ln{\sigma(x_i)} + \left(\frac{y_i-\mu(x_i)}{\sigma(x_i)}\right)^2}
\end{equation}
which amounts to minimizing the exponent
\begin{equation}\label{eq8}
\sum_{i=1}^N 2\ln(\sigma(x_i)) + \left(\frac{(y_i - \mu(x_i))}{\sigma(x_i)}\right)^2
\end{equation}
Note that when considering $\sigma$ is consent, minimizing $L_{uncert}$ amounts to minimizing $(y - \mu(x))^2$, the MSE.

In practice, when training the neural network using a batch size $B < N$ and over multiple biomass catergories $C$, we minimize

\begin{equation}\label{eq9}
L_{uncert} = \frac{1}{B}\sum_{i=1}^B\frac{1}{C}\sum_{c=1}^C 2\ln(\sigma_c(x_i)) + \left(\frac{(y_{i,c} - \mu_c(x_i))}{\sigma_c(x_i)}\right)^2
\end{equation}
also known as the variance attenuation loss~\cite{2022_CVPRW_varianceattenuation}.

Intuitively, if $(y_{i,c} - \mu_c(x_i))$ is close to zero, $L_{uncert}$ encourages the network to predict a small value for $\sigma_c(x_i)$ effectively minimizing the $2\ln(\sigma_c(x_i))$ term. However, if the prediction for $\mu_c(x_i)$ is inaccurate, a large $\sigma_c(x_i)$ is encouraged so that the $\left(\frac{(y_{i,c} - \mu_c(x_i))}{\sigma_c(x_i)}\right)^2$ fraction can be minimized  while the $2\ln\sigma_c(x_i)$ balances naive high $\sigma_c(x_i)$ solutions.

In practice, we predict $\mu(x_i)$ and $\ln\sigma(x_i)$ (for stability purposes) using two linear prediction heads sharing the same feature extractor, see Figure~\ref{Fig2}. At test time, we can evaluate the confidence of a given prediction by averaged spread of the predicted distributions, indicated by $\sigma(x_i) = \sum_{c=1}^C \sigma_c(x_i)$.


\textbf{Visualizing high variance images.}
Figure~\ref{irishvar} displays examples of low and high variance phone images predicted at test-time on the Irish clover dataset.  While the low variance images in the top row of Figure~\ref{irishvar} appears to be composed mostly of tall grass, higher variance images contain more clover. It is also interesting to notice that the detected high error images in the bottom row contain non uniform herbage height (left), many clover and clover flowers (all images) or foreign objects (piece of red rope in the rightmost image). 
As for the GrassClover dataset in Figure~\ref{fig:danishvar}, low variance images in the top row appear to contain mostly grass with low amounts of visible soil. High loss images in the bottom row however contain more clover and present a denser herbage with no visible soil.

For both datasets, clover-heavy images appear harder to correctly predict, especially for a pruned model. This problem is expectedly exacerbated for the GrassClover dataset that expects the algorithm to further differentiate between red and white clover species. It is also possible that the clover leaves in the herbage canopy grow above and occult a part of the grass underneath (middle and rightmost image in Figure~\ref{fig:danishvar}'s bottom row). In the rest of this paper, we refer to images where the pruned model outputs high error predictions as the hard images.

\textbf{Correlation between variance and RMSE.} Figure \ref{Fig3}(a) and Figure \ref{Fig4}(a) shows the correlation between the test RMSE and variance ($\sigma$) over different held-out phone images from the Irish dataset. We observe that the test time RMSE and variance are positively correlated with each other, even on images unseen during training. For example, VGG-16 in Figure \ref{Fig3}(a) exhibits a significant and positive pearson correlation of $0.71$. Given this information, we can re-infer high variance predictions using the unpruned model which yield a lower error. For example, as shown in Figure~\ref{Fig2}, initially all incoming images in a batch are inferred through pruned (energy-efficient) model. Then, for images where model's output variance does not fall under given threshold, they are re-inferred using original (unpruned) model. This approach circumvent the direct usage of the energy-intensive unpruned model for all images, effectively reducing energy consumption.  
\subsection{Hybridizing Models for Energy-efficient and Accurate Biomass Prediction}
To utilize the strengths of pruned and unpruned models jointly, we propose to leverage both settings by creating a hybrid system. Given a test image, we first predict biomass composition and variance from the pruned (energy-efficient) model, then if the variance exceeds a hyper-parameter threshold we re-infer using the unpruned (energy-intensive) model to obtain the final prediction. In essence, the threshold on the log-variance effectively acts as a control knob to balance the energy-accuracy trade-off in real time. Our hybrid approach is illustrated in Figure~\ref{Fig2}.

\section{Evaluation}
We conduct here experiments to quantify the effect of various pruning ratios $\epsilon$ on the regression accuracy of a state-of-the-art prediction algorithm~\cite{2022_CVPRW_IrishSuperRes}. We run experiments on the IrishClover and the GrassClover datasets using ResNet18~\cite{2016_CVPR_ResNet} and VGG16~\cite{2015_ICLR_VGG} architectures where although energy consumption diminishes, we observe an accuracy drop after aggressive pruning ($\epsilon > 50$). We then evaluate the performance of our pruned/unpruned hybrid algorithm on an edge device where we demonstrate that we can maintain accuracy while decreasing energy consumption from $40\%$ to $60\%$.

\begin{table}[]
 \caption{Results for pruned models, averaged over 5 runs $\pm$ std Irish dataset \label{tab:resultscompirish}. Unpruned results ($\varepsilon=0$) are obtained using publicly available code from~\cite{2022_CVPRW_IrishSuperRes}}
    \global\long\def\arraystretch{1}%
    \centering
    \resizebox{.4\textwidth}{!}{{{}}%
    \begin{tabular}{l>{\centering}c>{\centering}c>{\centering}c>{\centering}c>{\centering}c>{\centering}c>{\centering}c>{\centering}c}
    \toprule
        Network & $\varepsilon$ & RMSE Camera & RMSE Phone\tabularnewline
        \midrule
        \multirow{5}{*}{ResNet18} & 0 & $3.11\pm0.09$ & $4.81\pm0.68$\tabularnewline
         & 20 &  $4.38\pm0.19$ &  $5.55\pm1.32$ \tabularnewline
         & 40 & $4.44\pm0.05$ & $5.32\pm0.16$ \tabularnewline
         & 60 & $4.45\pm0.16$ & $5.75\pm1.02$ \tabularnewline
         & 80 & $4.44\pm0.10$ & $5.40\pm0.44$\tabularnewline
        \midrule
        \multirow{4}{*}{VGG16} & 0 & $3.84\pm0.12$ & $5.28\pm0.37$ \tabularnewline
         & 20 & $3.75\pm0.10$ & $6.82\pm1.57$ \tabularnewline
         & 60 & $3.84\pm0.22$ & $6.11\pm0.74$ \tabularnewline
         & 90 & $3.66\pm0.20$ &  $7.50\pm2.96$ \tabularnewline
        
       \bottomrule
    \end{tabular}}

\end{table}

\begin{table}[]
\caption{Results for pruned models, averaged over 5 runs $\pm$ std. GrassClover dataset \label{tab:resultscompdanish}. Unpruned results ($\varepsilon=0$) are obtained using publicly available code from~\cite{2022_CVPRW_IrishSuperRes}}
    \global\long\def\arraystretch{1}%
    \centering
    \resizebox{.25\textwidth}{!}{{{}}%
    \begin{tabular}{l>{\centering}c>{\centering}c>{\centering}c>{\centering}c>{\centering}c>{\centering}c>{\centering}c>{\centering}c}
    \toprule
        Network & $\varepsilon$ & RMSE Camera\tabularnewline
        \midrule
        \multirow{5}{*}{ResNet18} & 0 & $10.06\pm0.28$ \tabularnewline
         & 20 &  $10.91\pm0.42$ \tabularnewline
         & 40 & $11.25\pm0.37$  \tabularnewline
         & 60 & $11.28\pm0.35$ \tabularnewline
         & 80 & $11.21\pm0.10$ \tabularnewline
        \midrule
        \multirow{4}{*}{VGG16} & 0 & $11.83\pm0.32$  \tabularnewline
         & 20 & $13.16\pm0.42$  \tabularnewline
         & 60 & $13.67\pm1.10$  \tabularnewline
         & 90 & $13.52\pm0.80$  \tabularnewline
        
       \bottomrule
    \end{tabular}}

\end{table}

\subsection{Evaluation Settings}

We conduct biomass composition estimation experiments on the GrassClover~\cite{2019_CVPRW_grasscloverdataset} and Irish clover~\cite{2021_EGF_Irishdataset} datasets. We propose to train two different network architectures: ResNet18~\cite{2016_CVPR_ResNet} and VGG-16~\cite{2015_ICLR_VGG}. 
We run experiments for $100$ epochs and adopt a cosine learning rate decay. Since we do not need to estimate uncentainty for the unpruned models, we use the work of~\cite{2022_CVPRW_IrishSuperRes} as the uncompressed baseline without making use of any unlabeled images. The uncompressed baseline is trained using a Root Mean Square Error (RMSE) from a learning rate of $0.1$ for ResNet18 and $0.03$ for VGG16. For the pruned models, we train with $L_{uncert}$ and an initial learning rate of $0.0003$. For all experiments we train at a resolution of $224\times 224$ and utilize data augmentations that preserve all the information in the image: random horizontal and vertical flipping as well as random grayscaling with a probability of $0.2$. These training schedules follow previous implementations~\cite{2022_CVPRW_IrishSuperRes,2021_ICCVW_semisupmine,2022_EGF_Irishunsuppre}.

\subsection{Compression rate vs prediction accuracy}
We first study the effect of the compression rate $\varepsilon$ on the prediction error of the network. To do so, we generate five independent pruned models for each compression percentage $\varepsilon$. We then train the pruned networks using $L_{uncert}$ from equation~\ref{eq9} while the unpruned models are trained using a RMSE loss.
For the Irish dataset, we train the network on the  high quality camera images and evaluate on a held out set of camera or phone images to simulate edge conditions. For the GrassClover dataset, we report our results on the validation set which only contains high quality images. We repeat the experiments for VGG16 and ResNet18. Results are available in Tables~\ref{tab:resultscompirish} and ~\ref{tab:resultscompdanish}. We observe in the case of the Irish dataset that the pruned models manage to maintain good accuracy results on the higher quality camera images despite the compression but that the performance drops on the more challenging camera images. 
We observe a similar increase in RMSE when training pruned networks on the GrassClover dataset in Table~\ref{tab:resultscompdanish} even if all the images we test on are of high quality.

Interestingly, increasing $\varepsilon$ from $20\%$ to $90\%$ does not significantly increase the test error for ResNet18 or VGG16 on either dataset. These results may hint toward dry biomass composition being a task not requiring large amounts of parameters to solve, at least regarding high quality camera images. 
The accuracy difference between unpruned and pruned models, especially in the case of the phone images does however motivate a hybrid  unpruned-pruned approach.

\subsection{Energy efficiency and RMSE on edge}
We evaluate the energy efficiency and RMSE on a real edge device to compare the proposed approach with pruned-only and unpruned-only.

\textbf{Edge Device.}
We performed experiments for inference on a resource-constrained edge device i.e., NVIDIA Jetson Nano. This is a single-board device, similar to Raspberry Pi, but with a GPU to handle edge AI workloads. This edge device comprises a GPU with 128 NVIDIA CUDA® cores, a Quadcore ARM Cortex-A57 CPU, 4GB of RAM, and 16GB of microSD storage. 

\begin{figure}
\includegraphics[width=0.5\textwidth]
{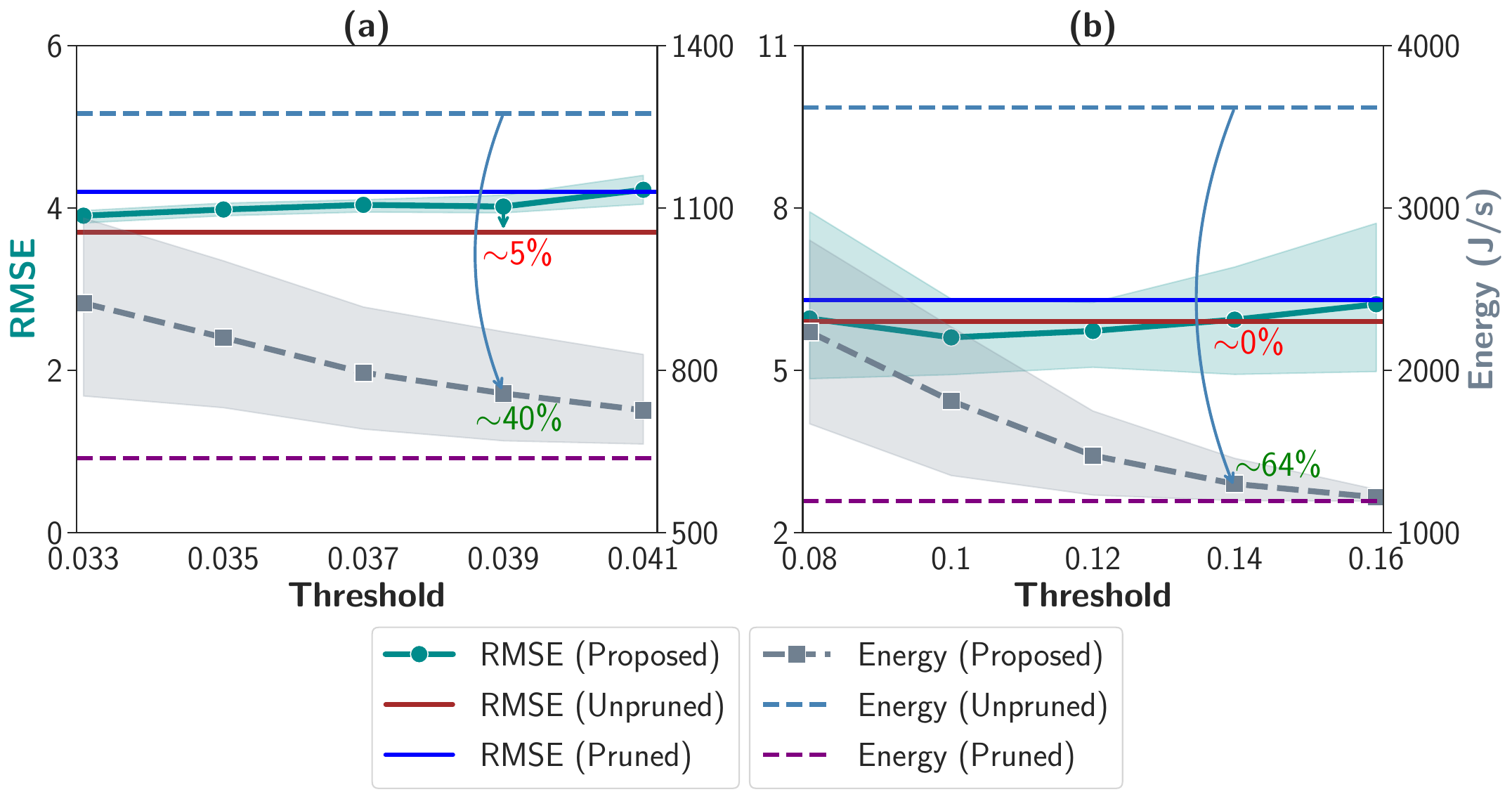}
\caption{Comparison of proposed approach with pruned-only and unpruned-only methods for energy efficiency and RMSE on Irish clover dataset, and (a) VGG-16, and (b) ResNet-18 models.}
\label{irish}
\end{figure}

\begin{figure}
\includegraphics[width=0.5\textwidth]
{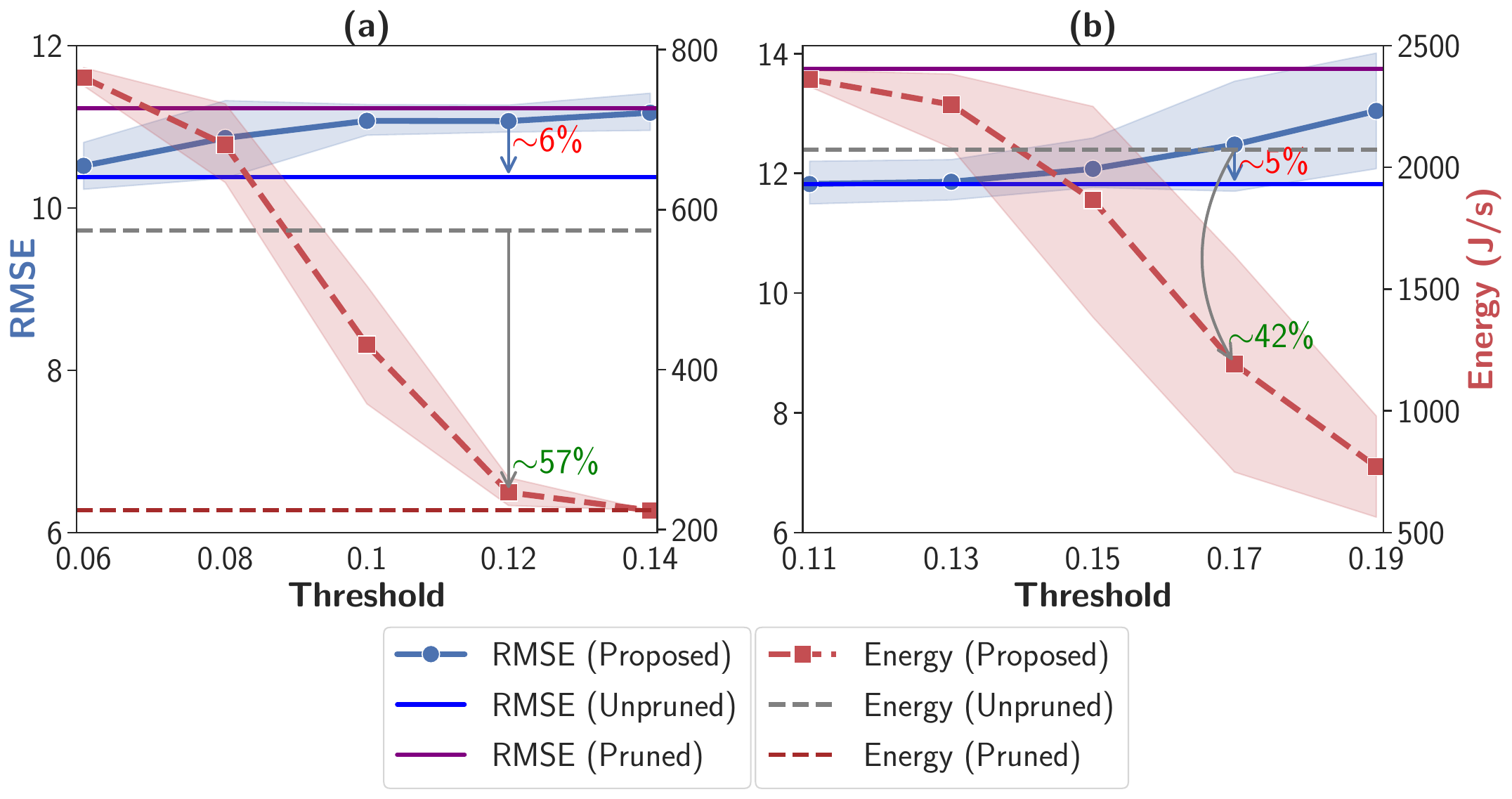}
\caption{Comparison of proposed approach with pruned-only and unpruned-only methods for energy efficiency and RMSE on GrassClover dataset, and (a) ResNet-18, and (b) VGG-16 models.}
\label{danish}
\end{figure}

\textbf{Proposed vs pruned / Unpruned.}
Figures \ref{danish} and \ref{irish} show how our approach can reduce energy consumption by $50\%$ on average with respect to the original unpruned model, while only losing an average of $4\%$ in accuracy. These experiments are repeated over 5 runs, and the mean results are presented with $95\%$ confidence interval.

In Figures \ref{danish}(a) and \ref{danish}(b), we show a performance evaluation over different threshold values for ResNet-18 and VGG-16 models respectively on the GrassClover dataset~\cite{2019_CVPRW_grasscloverdataset}. Similarly, in Figures \ref{irish}(a) and \ref{irish}(b), different threshold values are evaluated for VGG16 and ResNet18 models respectively on the Irish clover~\cite{2021_EGF_Irishdataset} dataset. In particular, we indicate the optimal threshold value for each model on both datasets to achieve high energy efficiency and low RMSE we achieve using our approach.  

\textbf{Number of re-inferred images.}
Table~\ref{tab:camnphone} shows the proportion of camera and phone images of the Irish dataset being refined by unpruned model according to different variance thresholds. We observe that phone images are re-inferred more often than the high resolution camera images. This indicates that the pruned algorithm struggles to generate to these harder images, unseen during training. Although the unpruned model was not seen phone images during training, it manages to generalize better that the pruned one, resulting in smaller prediction errors.


\section{Conclusion}
Evaluating the composition of dry biomass in grass provides valuable information for farmers. This assessment helps farmers enhance the quality of their grass and make informed decisions about local fertility improvement. Although computer vision algorithms based on deep learning have shown great potential in predicting accurate composition of dry biomass content, these algorithms are energy-intensive hence hampering their usage on resource-constrained edge devices. We first propose to apply filter pruning via energy-driven layer selection to reduce energy consumption. We however find that pruned model can suffer to generalize to challenging samples containing large amounts of clover as well as imperfect real world phone images. To retain accuracy while reducing energy usage, we propose to evaluate the uncertainty of the regression prediction using a variance attenuation loss. At test-time, only harder images whose predicted variance exceeds a threshold are refined by the more accurate unpruned network. We observe that this hybrid approach only slightly increases the error rate of previous algorithms while largely reducing the energy requirements of deep learning biomass estimation algorithms. 
Finally, generating a variance metric for the network's predictions, which correlate with the actual biomass prediction error, is crucial for encouraging acceptance among human experts. In practical scenarios, identifying a high predicted variance would prompt end users to either take a clearer picture of the grass or to approach the network's predictions with due caution.

\begin{table}[t!]
 \caption{Number of Camera and Phone images being re-inferred by unpruned model on Irish dataset. Each validation image has a phone and a camera view for a total of $214$ images ($107$ in each view)~\label{tab:camnphone}}
    \global\long\def\arraystretch{1}%
    \centering
    \resizebox{.4\textwidth}{!}{{{}}%
    \begin{tabular}{l>{\centering}c>{\centering}c>{\centering}c>{\centering}c>{\centering}c>{\centering}c>{\centering}c>{\centering}c}
    \toprule
        Network & Threshold & \# of Camera  & \# of Phone & Total\tabularnewline
        \midrule
        \multirow{5}{*}{ResNet18} & 0.033 & $16$ & $32$ & $48$\tabularnewline
         & 0.035 & $9$ & $18$ & $27$ \tabularnewline
         & 0.037 & $9$ & $17$ & $26$ \tabularnewline
         & 0.039 & $6$ & $13$ & $19$ \tabularnewline
         & 0.041 & $4$ & $11$ & $15$\tabularnewline
        \midrule
        \multirow{4}{*}{VGG16} & 0.08 & $28$ & $32$ & $60$ \tabularnewline
         & 0.10 & $20$ & $15$ & $35$ \tabularnewline
         & 0.12 & $8$ & $9$ & $17$ \tabularnewline
         & 0.14 & $2$ &  $5$ & $7$  \tabularnewline
         & 0.16 & $0$ &  $2$ & $2$ \tabularnewline
        
       \bottomrule
    \end{tabular}}

\end{table}

\section*{Acknowledgment}

This work  was  supported  by the Science  Foundation  Ireland (SFI), by the  Department  of  Agriculture,  Food, and Marine on behalf of the Government of Ireland VistaMilk research centre under award 16/RC/3835, and in part by SFI under Grant 21/FFP-A/9174, by NGI Enrichers program under Horizon Europe grant: 101070125, and by the National Science Foundation (NSF) under the CPS award~\#~2133407.

{
    \small
    \bibliographystyle{ieeenat_fullname}
    \bibliography{main}
}


\end{document}